%% file: paper.tex
\documentclass{article}

\input{math_commands.tex}

\usepackage{microtype}
\usepackage{graphicx}
\usepackage{subfigure}
\usepackage{wrapfig}
\usepackage{url}
\usepackage{booktabs, multirow} 
\usepackage{soul}
\usepackage{amsthm}
\usepackage[section]{placeins}

\usepackage{hyperref}

\usepackage[accepted]{icml2023}

\usepackage{amsmath}
\usepackage{amssymb}
\usepackage{mathtools}
\usepackage{amsthm}

\usepackage[capitalize,noabbrev]{cleveref}

\theoremstyle{plain}

\theoremstyle{definition}

\theoremstyle{remark}

\newtheorem*{defn}{Definition}

\usepackage[textsize=tiny]{todonotes}

\icmltitlerunning{Scaling Laws for Multilingual Neural Machine Translation}

\newcommand{\bp}{\bm{p}}
\begin{document}

\twocolumn[
\icmltitle{
Scaling Laws for Multilingual Neural Machine Translation
}



\icmlsetsymbol{equal}{*}

\begin{icmlauthorlist}
\icmlauthor{Patrick Fernandes}{google,cmu,ist}
\icmlauthor{Behrooz Ghorbani}{google}
\icmlauthor{Xavier Garcia}{google}
\icmlauthor{Markus Freitag}{google}
\icmlauthor{Orhan Firat}{google}
\end{icmlauthorlist}

\icmlaffiliation{google}{Google Research}
\icmlaffiliation{cmu}{Carnegie Mellon University}
\icmlaffiliation{ist}{Instituto Superior Técnico}

\icmlcorrespondingauthor{Patrick Fernandes}{pfernand@cs.cmu.edu}

\icmlkeywords{Scaling Laws, Multilingual Translation, Neural Machine Translation, Multitask Models, ICML}

\vskip 0.3in
]



\printAffiliationsAndNotice{}  

\begin{abstract}
In this work, we provide a large-scale empirical study of the scaling properties of multilingual neural machine translation models. We examine how increases in the model size affect the model performance and investigate the role of the training mixture composition on the scaling behavior. We find that changing the weightings of the individual language pairs in the training mixture only affect the multiplicative factor of the scaling law. In particular, we observe that multilingual models trained using different mixing rates all exhibit the same scaling exponent. Through a novel joint scaling law formulation, we compute the \textit{effective number of parameters} allocated to each language pair and examine the role of language similarity in the scaling behavior of our models. We find little evidence that language similarity has any impact. In contrast, the direction of the multilinguality plays a significant role, with models translating from multiple languages \textit{into} English having a larger number of effective parameters per task than their reversed counterparts. Finally, we leverage our observations to predict the performance of multilingual models trained with \textit{any} language weighting at \textit{any} scale, significantly reducing efforts required for language balancing in large multilingual models. Our findings apply to both in-domain and out-of-domain test sets and to multiple evaluation metrics, such as ChrF and BLEURT. \looseness=-1
\end{abstract}

\section{Introduction}
Over the past few years, scaling has emerged as a popular and effective way to improve the performance of neural networks \citep{gpt3, palm, lepikhin2020gshard}. Given the costs associated with training large neural models, much work has gone into understanding their scaling properties and predicting the evolution of their performance with scale through \textbf{scaling laws}. Such scaling laws have been instrumental in guiding the model development efforts across a variety of domains such as computer vision \citep{zhai2022scaling}, language modeling \citep{scaling-laws-lm, chinchilla}, and neural machine translation \citep{scaling-laws-mt}. \looseness=-1

Despite these impressive developments, most of the scaling laws studies available in the literature only focus on single-task, single-language models. On the contrary, current massive neural models are often trained to solve more than one task across one or more modalities \& languages \citep{palm, sanh2022multitask, reed2022generalist}. This disconnect from the current research frontier limits the applicability of scaling laws in guiding model development decisions. In particular, currently available scaling laws studies are unable to inform the decision process on \textbf{balancing the different tasks effectively} at training time. Without such guidance, practitioners often have to rely on cumbersome and costly approaches such as approximate grid search to inform their decision-making; such approaches quickly become infeasible as the problem scale grows.

In this paper, we take the initial step towards developing a quantitative understanding of the scaling behavior for multitask models. We choose multilingual neural machine translation (MNMT) as the setup for this initial study. This choice is motivated by several reasons: (1) MNMT has been framed and studied as a multi-task optimization problem extensively in the past \citep{dong-etal-2015-multi, luong-multitask, arivazhagan-mnmt-wild, wang2021gradient}; (2) It provides a popular setup with mature benchmarks and substantial literature on scaling \citep{lepikhin2020gshard, costa2022no, google-mt-ntl, huang2019gpipe}; (3) Moreover, recent results on scaling laws for single-task MT models provide a natural starting point for our study \citep{scaling-laws-mt, Bansal2022DataSL, gordon-etal-2021-data, zhang2022examining}. (4) Finally, recent findings on the optimization dynamics of MNMT models greatly simplify our study by removing the need to examine the role of the optimization algorithm in our results \citep{xin2022multitask}.

For our analysis, we train over $200$ MNMT models (ranging from $20$M to $1$B non-embedding parameters) and systematically examine their scaling behaviors. We focus our investigation on the \textbf{data-rich, compute-rich regime} where we have access to vast amounts of training data for all the language pairs (i.e. tasks)\footnote{Using machine translation terminology, all language pairs are \emph{high-resource}.} and the models are trained to near convergence. Here, the main bottleneck in the model performance is the lack of model capacity. We establish the following observations:
\begin{itemize}
\item For each fixed training mixture, the evolution of the test cross-entropy loss for the $i_{\textit{th}}$ language pair ($\Ls_i$) with model size ($N$) follows a scaling law that resembles the scaling behavior of single-language-pair models:
\begin{align}
    \Ls_i(N; \bp) \approx \beta_{\bp, i} N^{-\alpha_{\bp, i}} + L_{\infty}^{(\bp, i)}.
\end{align} 
Here, $\bp$ is a vector of probabilities that determines the weight of each language pair in the training mixture. Furthermore, we find that changes in the language pair weightings only affect the multiplicative factor $\beta$; the scaling exponent $\alpha$ and the irreducible loss $L_{\infty}$ are unaffected by these changes. As such, our results suggest that scaling multilingual models improves the loss at rates that are independent of the weights of the individual language pairs in the training mixture. 
%
\item We leverage these findings to propose a scaling law that jointly predicts the performance for all language pairs and weightings considered, and use it to examine how the model splits its capacity in between the language pairs by computing the \textbf{effective number of parameters} allocated to each language pair (Section \ref{subsec:capacity}). \looseness=-1

\item We examine the popular belief that training multilingual models on similar languages is more effective than training models in unrelated languages. Surprisingly, for the high-resource language pairs considered, we do not observe any significant differences in the scaling behavior of models trained to translate from English into related languages (En$\rightarrow$\{De, Fr\}) and models trained in unrelated languages (En$\rightarrow$\{De, Zh\}). In contrast, we observe that models trained to translate from multiple languages into English (XX$\rightarrow$En) benefit much more from multitasking compared to those trained on translation out of English (En$\rightarrow$XX).

\item By approximating the capacity splitting behavior of multilingual models, in Section \ref{subsec:task_balancing}, we provide a scaling law that predicts \textbf{the full task performance trade-off frontier} as a function of the model size $N$ (See Figure \ref{fig:ratio_and_pareto}). In Section \ref{subsec:task_balancing}, we describe how such predictions can be leveraged for efficiently guiding task balancing when training large multilingual models. 

\end{itemize}

\section{Background}
\subsection{Neural Scaling Laws}
Recent research suggests that the performance of large neural models is well-predicted by a smooth function of the fundamental problem parameters: the model size $N$ \footnote{Following the literature conventions, we only consider the non-embedding layers when computing $N$.}, the size of the training data $D$, and the amount of compute used for training $C$ \citep{hestness2017deep, rosenfeld2019constructive, scaling-laws-lm, Hernandez2021ScalingLF}. The most relevant of these studies to ours is \citet{scaling-laws-mt} where the authors study the effects of increasing the model size for single-task NMT models in the data-rich ($D\rightarrow \infty$), compute-rich ($C\rightarrow \infty$) regime. In this setting, the authors show that the following \emph{bivariate} law describes the scaling behavior of encoder-decoder Transformers
\begin{align}
\label{eq:scaling-law-end-dec}
\Ls(N_e, N_d) = \beta N_e^{-\alpha_e} N_d^{-\alpha_d} + L_{\infty}.
\end{align}
Here, $N_e$ and $N_d$ correspond to the number of parameters in the encoder and decoder respectively and $L_{\infty}$ corresponds to the irreducible loss associated with the task. $\{\beta, \alpha_e, \alpha_d, L_{\infty}\}$ are the parameters of the scaling law that need to be empirically estimated from the data.

In addition, \cite{scaling-laws-mt} examine the question of optimally allocating parameters between the encoder and the decoder. They show that in order to attain the optimal scaling behavior, one needs to proportionally scale the encoder and the decoder together. Under such scaling scheme,  \autoref{eq:scaling-law-end-dec} simplifies to 
\begin{align}
\label{eq:scaling-law-opt}
\Ls(N) = \beta N^{-\alpha} + L_{\infty},
\end{align}
which is similar to the scaling behavior observed in other domains such as computer vision \citep{zhai2022scaling} and autoregressive generative models \citep{henighan2020scaling}. 

Based on these results, to achieve the optimal scaling behavior, we adopt the proportional encoder-decoder scaling scheme for our experiments. A detailed overview of the size and architecture of our models is presented in Appendix \ref{app:model-details}.

\subsection{Multitask Optimization}

Multilingual NMT is commonly framed and studied as a multitask optimization problem \citep{dong-etal-2015-multi, luong-multitask, arivazhagan-mnmt-wild, wang2021gradient}.

We focus our investigation on the supervised learning setup where the model parameters $\vtheta \in \R^N$ are trained on $K$ different tasks simultaneously. In multilingual MT, each task corresponds to translation for a different language pair. We denote the loss associated with task $i$ with $\Ls_{i}(\vtheta)$. 

Multitask models are often trained by minimizing a convex combination of the per-task losses:
\begin{align} \label{eq:scalarization}
\hat{\vtheta}(\vw) = \argmin\sum_{i=1}^K \vw_i \Ls_i(\vtheta) \; ; \; \vw > 0, \; \sum_{i=0}^K \vw_i = 1
\end{align}
Here, $\vw$ is a fixed vector of the task weights, determined apriori by the practitioner to emphasize her preferences on the balancing of the tasks. This so-called \textbf{scalarization} approach is highly popular in the community due to its effectiveness and simplicity.\footnote{See \citep{boyd_vandenberghe_2004} for more a detailed discussion of scalarization.} In fact, despite this simplicity, recent results on multitask optimization suggest that scalarization achieves performances on par or better than bespoke optimizers designed specifically for multitask models \citep{xin2022multitask, kurin2022defense}. 

In current large text models, such explicit scalarization is rare. Instead, scalarization is often implemented \textbf{implicitly}, by sampling observations from each task proportionally to that task's weight. Proportional sampling produces (in expectation) the same overall loss function as explicit scalarization but with much less engineering complexity. 

\begin{figure}
\centering
\includegraphics[width=0.7\columnwidth]{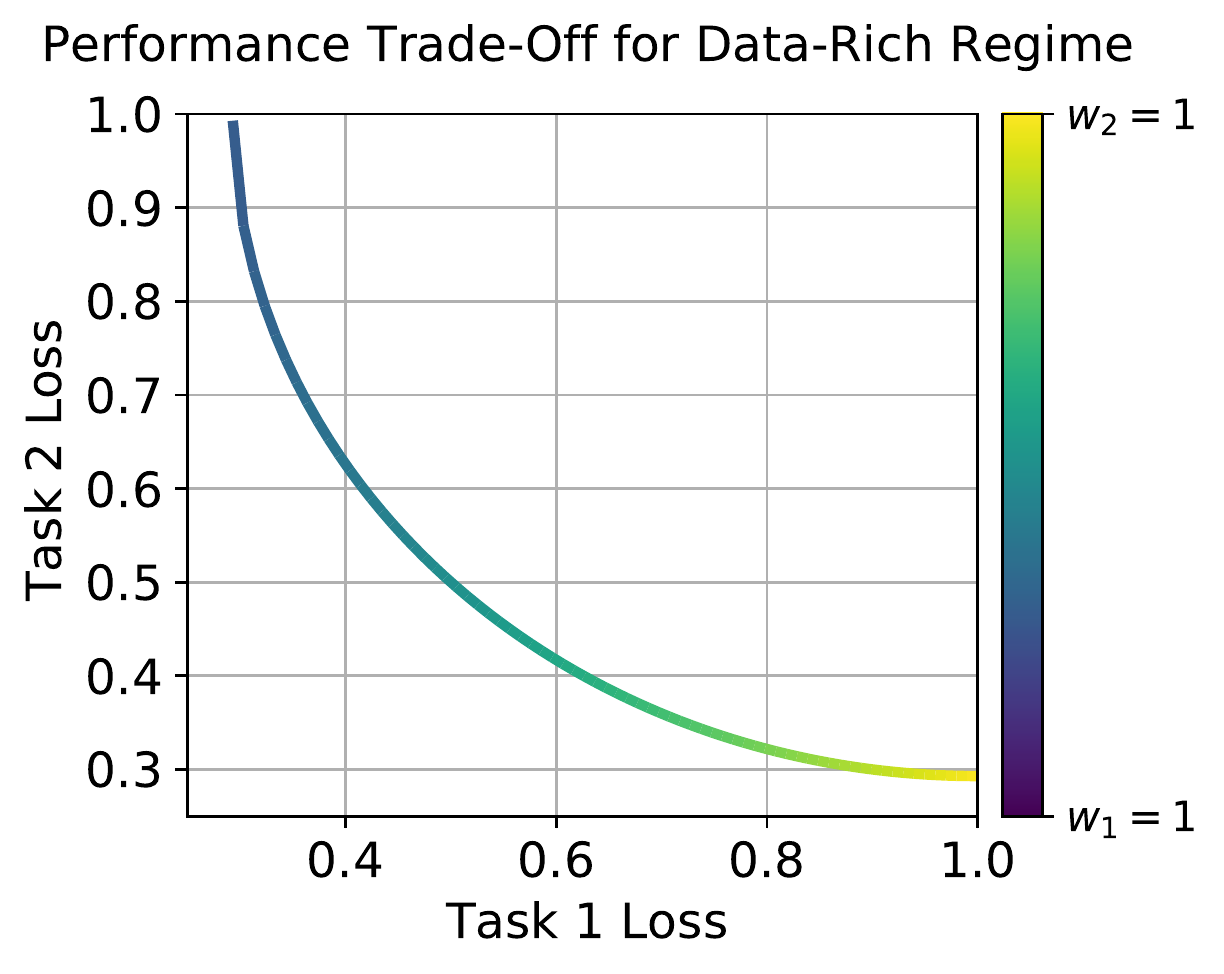}
\vspace{-0.1cm}
\caption{\label{fig:toy-multitask} Cartoon representation of the performance trade-off frontier for a hypothetical model.}
\vspace{-0.5cm}
\end{figure}

\citet{xin2022multitask} demonstrate that there exists a smooth, well-defined performance trade-off frontier for multitask models in the data rich regime. This frontier represents the performance trade-off the model is able to achieve in between the tasks as a function of the task weights (see Figure \ref{fig:toy-multitask} for a cartoon representation). Naturally, finding an accurate characterization of the performance trade-off frontier is key in finding a systematic solution to the task balancing issue.

\section{Effects of Scale in Multilingual MT}

\subsection{Experimental Setup}
\label{sec:experimental-setup}
We use the pre-LN encoder-decoder Transformer architecture in our models \citep{xiong2020layer, vaswani2017attention}. We train models of up to $8$ sizes, approximately ranging from $20$M to $1$B (non-embedding) parameters. When scaling encoder-decoder Transformers, to achieve the optimal scaling behavior, we scale the encoder and the decoder proportionally by increasing the model dimension and the number of layers in tandem. See \autoref{app:model-details} for details.

For our experiments, we train two cohorts of models: En$\rightarrow$XX and XX$\rightarrow$En. For En$\rightarrow$XX cohort, we train multilingual model for translation from English to \{German (De), Chinese (Zh)\} and \{German (De), French (Fr)\}. For XX$\rightarrow$En cohort, we present results for \{De, Zh\}$\rightarrow$En. 

We use the \textit{implicit} scalarization approach to train our models; each observation in the training batch is chosen from the first language pair with probability $p$ and the second language pair with probability $1-p$. For our experiments, we choose $p$ from the set 
\begin{align}
\label{eq:p-values}
p \in \{0, 0.05, 0.1, 0.3, 0.5, 0.7, 0.9, 0.95, 1\}.
\end{align}

For En$\rightarrow$XX models, to avoid confusing the model, we prepend a language token to the source sentence specifying the target language (e.g. \texttt{<2de>}).
The models are trained with per-token cross-entropy loss and Adafactor optimizer \citep{Shazeer2018AdafactorAL}, using a fixed batch size of $500$K tokens and inverse square root learning rate schedule. To mirror the compute-rich regime as closely as possible, we trained our models to near convergence. In practice, this translates to training our smaller models ($<500$M parameters) for 500K gradient steps and larger models for $1$M steps. \looseness=-1

To place our models in the data-rich regime, we use a massive in-house web-crawled dataset for training our models.
We filter this data using an online data selection procedure \citep{wang-etal-2018-denoising} and high-quality web-domain reference sets, extracting 600M sentences for each language pair in the En$\rightarrow$XX direction and 1.2B sentences for the XX$\rightarrow$En language pairs. We tokenize this corpus by using a pretrained multilingual SentencePiece \cite{kudo-2018-subword} vocabulary, with a size of $128$K sub-words.

We measure the performance of models on both \textit{in-domain} and \textit{out-of-domain} test sets. For the in-domain test set, we extract $2000$ sentences from the same in-house datasets used to create the training (ensuring no overlap). For out-of-domain, we use \textit{newstest2019} \citep{barrault-etal-2019-findings}, consisting of $2000$ sentence-pairs extracted from aligned news documents. 

\subsection{Results \& Analysis}
\label{sec:experimental-setup}
\paragraph{Understanding Multilingual Scaling} 

\begin{figure*}[h]
\centering
\includegraphics[width=0.85\textwidth]{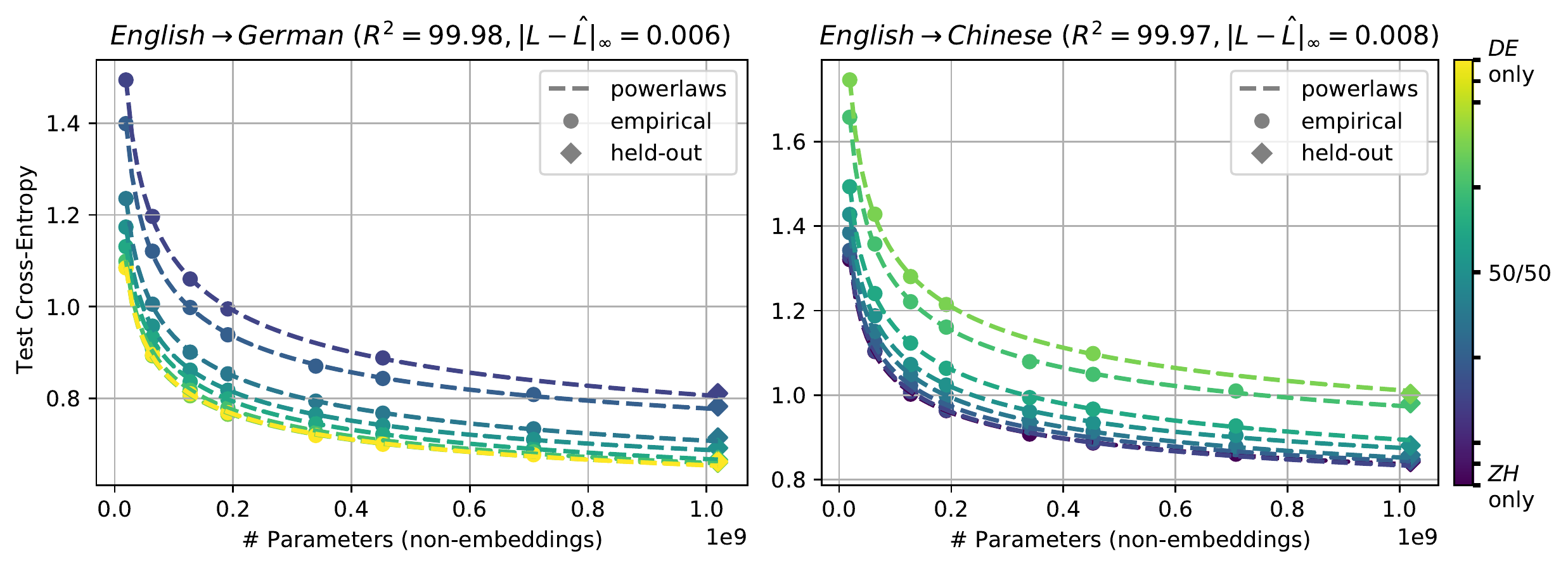}
\caption{\label{fig:individual-sl-en-dezh} The evolution of the in-domain test cross-entropy loss with model size for En$\rightarrow$\{De, Zh\} models, as well as the fitted scaling laws. These scaling laws are \textbf{fitted separately for each language pair weighting}. The color represents the weighting of the languages. The scaling laws are able to capture close to 100\% of the variation in the data for both language pairs. Note that we don't show the \textit{zero-shot} behavior.}
\end{figure*}

\begin{figure*}[h]
\centering
\includegraphics[width=0.85\textwidth]{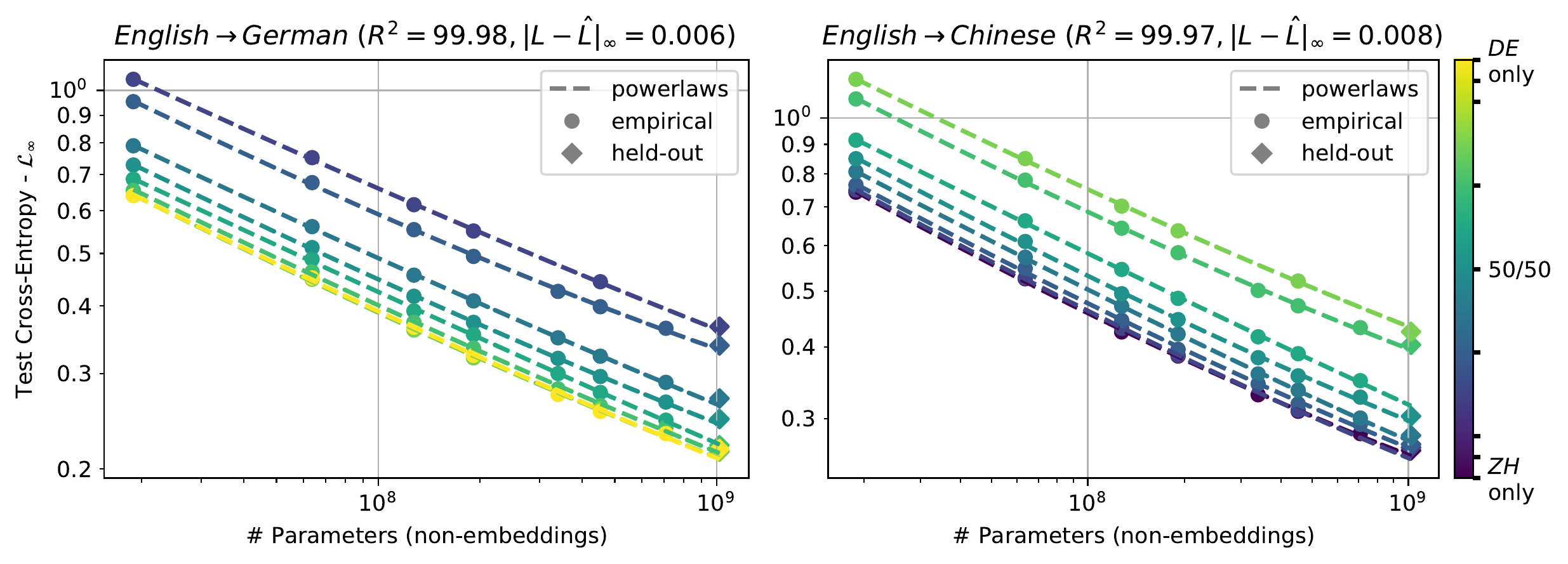}
\caption{\label{fig:individual-sl-en-dezh-log} Log-log plot of the evolution of the (in-domain) test cross-entropy loss as we scale. We subtract a constant {$L^{(i)}_\infty$}, jointly fitted for all the weightings (\autoref{eq:joint-scaling-law}). All lines are nearly parallel, suggesting that the scaling exponent is unchanged for all $p$.}
\end{figure*}

We start our analysis by independently examining the model scaling behavior for each individual language pair weighting $p$ in (\ref{eq:p-values}). For each choice of $p$, we fit a scaling law of the form
\begin{align} \label{eq:per_task_scaling}
    \Ls_i(N; p) = \beta_{p, i} N^{-\alpha_{p, i}} + L_{\infty}^{(p, i)}
\end{align}
 to the empirical (test) performance of models resulting from that language pair weighting.

\autoref{fig:individual-sl-en-dezh} presents our findings for En$\rightarrow$\{De, Zh\} models. Each point on the graph corresponds to the empirical test-cross entropy performance of a model at the end of the training.\footnote{For low probability language pairs, we apply a convergence correction procedure to make up for slow convergence. See \autoref{app:convergence-correction} for more details.} We observe that our per-weighting laws are able to capture the scaling behavior of our multilingual models on both language pairs. As expected, when the weight for one of the languages is decreased, the performance of the models on that language decreases for all scales. Our results suggest that the benefits of the increased model size for MNMT models are well-described by a power-law. See \autoref{app:individual-scaling-law-fits} for similar results for other language pair combinations.

\begin{figure*}[h]
\centering
\includegraphics[width=0.85\textwidth]{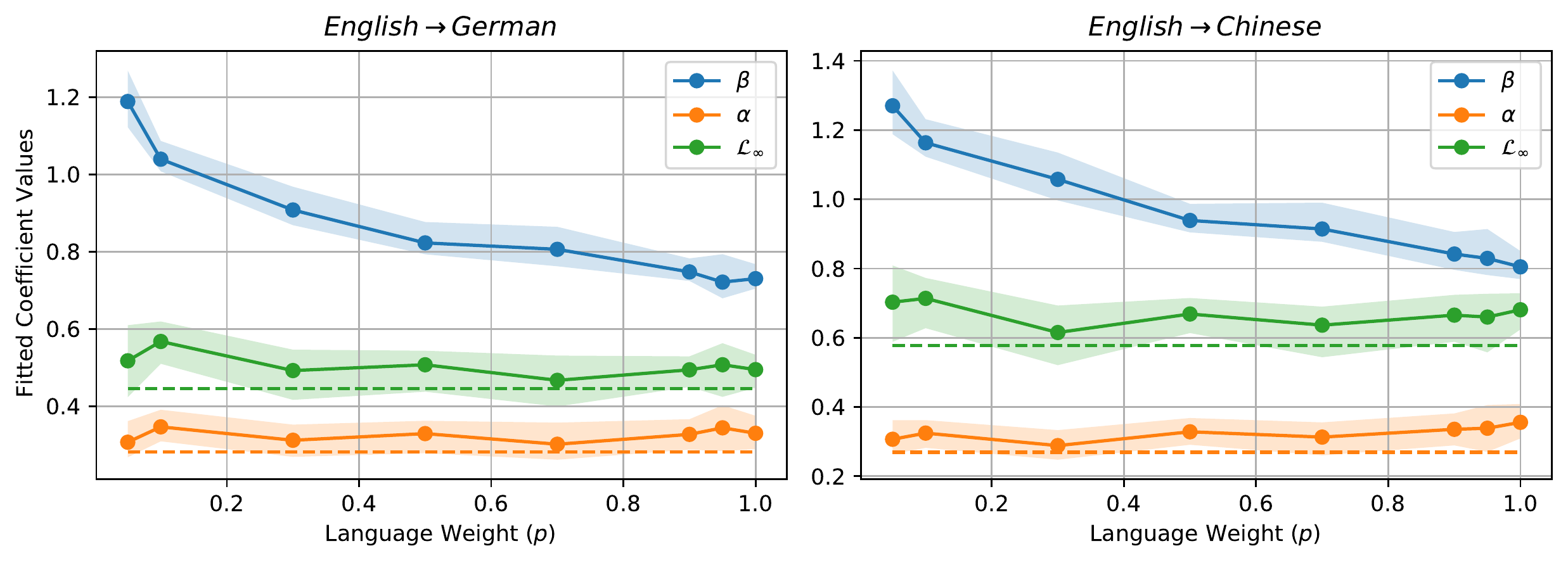}
\caption{\label{fig:scaling_coefficients_en_dezh}Coefficient values for German (left) and Chinese (right) as a function of the language weight, with the shaded region representing the standard deviation. The dashed lines represent the value of jointly fitted coefficients from \autoref{eq:joint-scaling-law}.}
\end{figure*}

\autoref{fig:scaling_coefficients_en_dezh} shows the fitted scaling law coefficients for different values of $p$. The shaded area marks the one standard deviation uncertainty interval of our estimates.\footnote{We gauge the uncertainty in the coefficients by measuring the fluctuations in our estimates when our empirical datapoints are perturbed by $\eps \overset{\text{i.i.d}}{\sim} \mathcal{N}(0, \sigma^2)$. We choose a conservative $\sigma$ of $1\%$ of the observed empirical loss for each data point.} Interestingly, we find that, across all values of $p$, both the scaling exponent ($\alpha$) and the irreducible loss ($\Ls_\infty$) seem to be relatively unchanged. In particular, all of our estimated $\alpha$ and $\Ls_{\infty}$ parameters are within two standard deviations of each other. In contrast, the multiplicative factor $\beta$ seems to be highly sensitive to the choice of $p$. 

\begin{figure*}[h]
\centering
\includegraphics[width=0.85\textwidth]{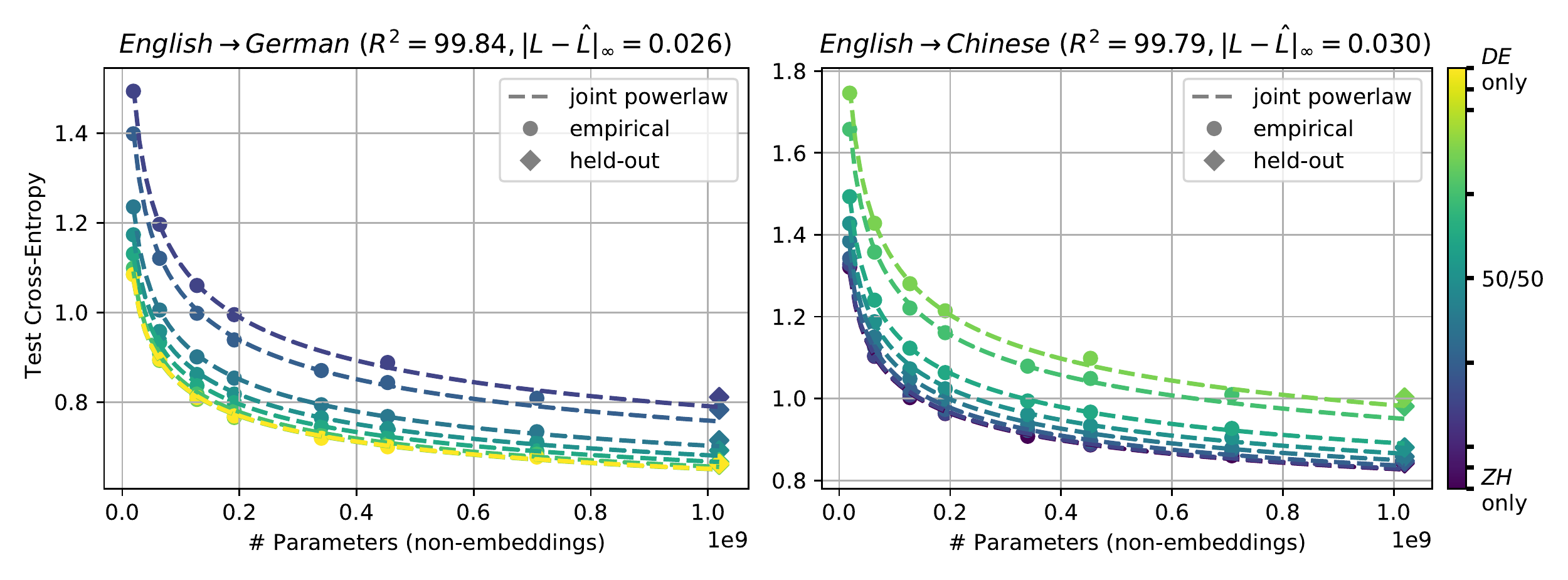}
\caption{\label{fig:joint-sl-en-dezh} The \textbf{joint} scaling law of \autoref{eq:joint-scaling-law} closely captures the scaling behavior of En$\rightarrow$\{De, Zh\} models. Test loss here is evaluated on in-domain test sets. See \autoref{app:joint-scaling-law-fits} for similar observations on En$\rightarrow$\{De, Fr\} and \{De, Zh\}$\rightarrow$En models.}
\end{figure*}

Figure \ref{fig:individual-sl-en-dezh-log} visually confirms the assertion that for our models $\alpha_p$ and $L_\infty$ are effectively constant. Here, we have subtracted a fixed constant $L^{(i)}_{\infty}$ from all the \autoref{fig:scaling_coefficients_en_dezh} curves corresponding to the language pair $i$. We then plot results on log-log axes. As the figure suggests, the lines are all near parallel, suggesting that the scaling exponent is unchanged for all $p$. In practical terms this means that, for example, doubling the capacity of a multilingual model will reduce its loss by the same $\frac{1}{2^\alpha}$ factor, no matter how the training mixture looks like. This also means that single-language-pair scaling laws can be used to gauge the benefits of scaling multilingual models.

\paragraph{Jointly Modeling Multilingual Scaling}

Based on the findings above, we make the assumption that the scaling exponents and the irreducible losses are independent of the language pair weights, and propose a \textbf{joint} scaling law of the form
\begin{align}
    \label{eq:joint-scaling-law}
    \Ls_i(N; p) \approx \beta_{p, i} N^{-\alpha_{i}} + L_{\infty}^{(i)}. 
\end{align} 

\autoref{fig:joint-sl-en-dezh} shows the fit of this joint scaling law for En$\rightarrow$\{De, Zh\} models evaluated on the in-domain test sets. Note that here, we fit a total of $10$ parameters for each language pair -- $8$ for $\beta_{p, i}$'s and two for $\alpha_i$ and $L_{\infty}^{(i)}$. In contrast, in Figure \ref{fig:individual-sl-en-dezh}, we used $24$ overall parameters to capture the scaling behavior for each language pair. Despite this significant decrease in the number of total fitted parameters, we observe that our joint laws are able to almost completely capture the scaling behavior. We observe a similar phenomenon for out-of-domain test sets and other language pairs (see \autoref{app:joint-scaling-law-fits}), further suggesting that the joint law accurately describes the scaling behavior of MNMT models.

\begin{figure*}[h]
\centering
\includegraphics[width=0.85\textwidth]{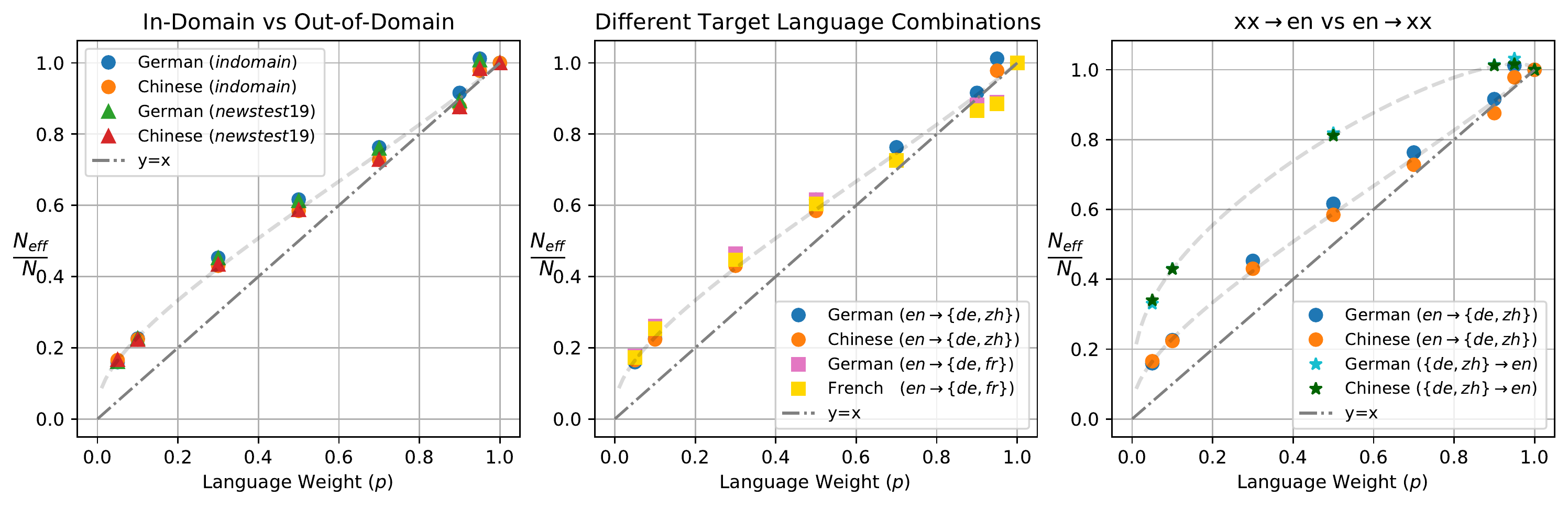}
\caption{\label{fig:ratios-comparion} The effective fraction of parameters allocated to each language pair as estimated by our joint scaling laws. Gray dashed lines correspond to the fitted $\hat{f}_i$ described in Equation \ref{eq:f_hat}. \emph{Left:} Comparison of the capacity splitting behavior of En$\rightarrow$\{De, Zh\} models for in-domain and out-of-domain test sets. We observe minimal differences between the two. \emph{Center:} Comparison of the capacity splitting behavior for En$\rightarrow$\{De, Zh\} and En$\rightarrow$\{De, Fr\} models. We don't observe any changes in the interaction between the language pairs based on language similarity. \emph{Right:} Comparison of the capacity splitting behavior for translation to and from English. XX$\rightarrow$En models exhibit more synergy among the language pairs.}
\end{figure*}

\subsection{Effective Network Capacity for Multilingual Models} \label{subsec:capacity}
We leverage our joint scaling law to examine how MNMT models split their capacity in between the different language pairs. We start by defining the notion of \textbf{the effective number of parameters}:
\begin{defn} \label{def:n_eff}
Consider a multilingual model in which a language pair $i$ has been trained with weight $p$. We define the effective number of parameters allocated to $i$, $N^{(i, p)}_{\text{eff}}$, to be equal to the number of parameters necessary for a single-language-pair model solely trained on $i$ to reach the same (test loss) performance as the multilingual model. 
\end{defn}

Mathematically, $N^{(i, p)}_{\text{eff}}$ can be written as the solution of the equation
\begin{align}
    \Ls_i(N; p) = \Ls_i(N^{(i, p)}_{\text{eff}}; 1). 
\end{align} 
A simple derivation yields that \footnote{See \autoref{app:model-ratio-derivation} for details.}
\begin{align}
    \label{eq:effective_param}
    N^{(i, p)}_{\text{eff}} = \left ( \frac{\beta_{1, i}}{\beta_{p, i}} \right)^{\frac{1}{\alpha_{i}}} N.
\end{align} 
Crucially, our calculations suggest that the fraction of parameters allocated to language pair $i$, which we denote by $f_i(p)$, is independent of the model size:
\begin{align}
    \label{eq:parameter-ratio}
    f_i(p) \equiv N^{(i, p)}_{\text{eff}} / N = \left ( \frac{\beta_{1, i}}{\beta_{p, i}} \right)^{\frac{1}{\alpha_{i}}}.
\end{align} 

This observation yields a fundamental, scale-independent quantity that can be leveraged for understanding the interactions between the different language pairs in the model. 

Figure \ref{fig:ratios-comparion} shows the empirically estimated effective parameter ratios for our models. Several observations are in order:

\textbf{Consistency Across Domains:} In Figure \ref{fig:ratios-comparion} (left), we compare the capacity splitting behavior of the models on in-domain and out-of-domain (newstest19) test sets. Even though the scaling laws coefficients for in-domain and out-of-domain test sets differ, we observe that the capacity splitting behavior is mostly unchanged with different test sets. These findings hint at some measure of universality across test domains on how MNMT models divide their capacity and share their parameters.

\textbf{Consistency Across Languages Pairs:} In Figure \ref{fig:ratios-comparion} (center), we compare the capacity splitting behavior of En$\rightarrow$\{De, Zh\} and En$\rightarrow$\{De, Fr\} models. The conventional wisdom in the MT literature suggests that the tasks in En$\rightarrow$\{De, Fr\} should exhibit a more positive interaction with each other
compared to En$\rightarrow$\{De, Zh\}. This is often justified by the intuition that representations are more aligned in related languages and more aligned representations will encourage parameter sharing \citep{dabre-etal-2017-enabling}. Surprisingly, our results suggest that the interaction dynamics in En$\rightarrow$\{De, Fr\} and En$\rightarrow$\{De, Zh\} models are not significantly different. In both settings, we observe a relatively neutral multilingual behavior -- the performance of an MNMT model of size $N$ trained on language pair $i$ with (sampling) weight $p$ is essentially similar to a single-language-pair model of size $pN$. In other words, there is minimal synergy among the languages in both setups. 

\textbf{En$\rightarrow$XX vs XX$\rightarrow$En:} In Figure \ref{fig:ratios-comparion} (right), we compare the interaction between the language pairs when translating out of English vs when translating into English. 

\begin{figure*}[h]
\centering
\includegraphics[width=0.78\textwidth]{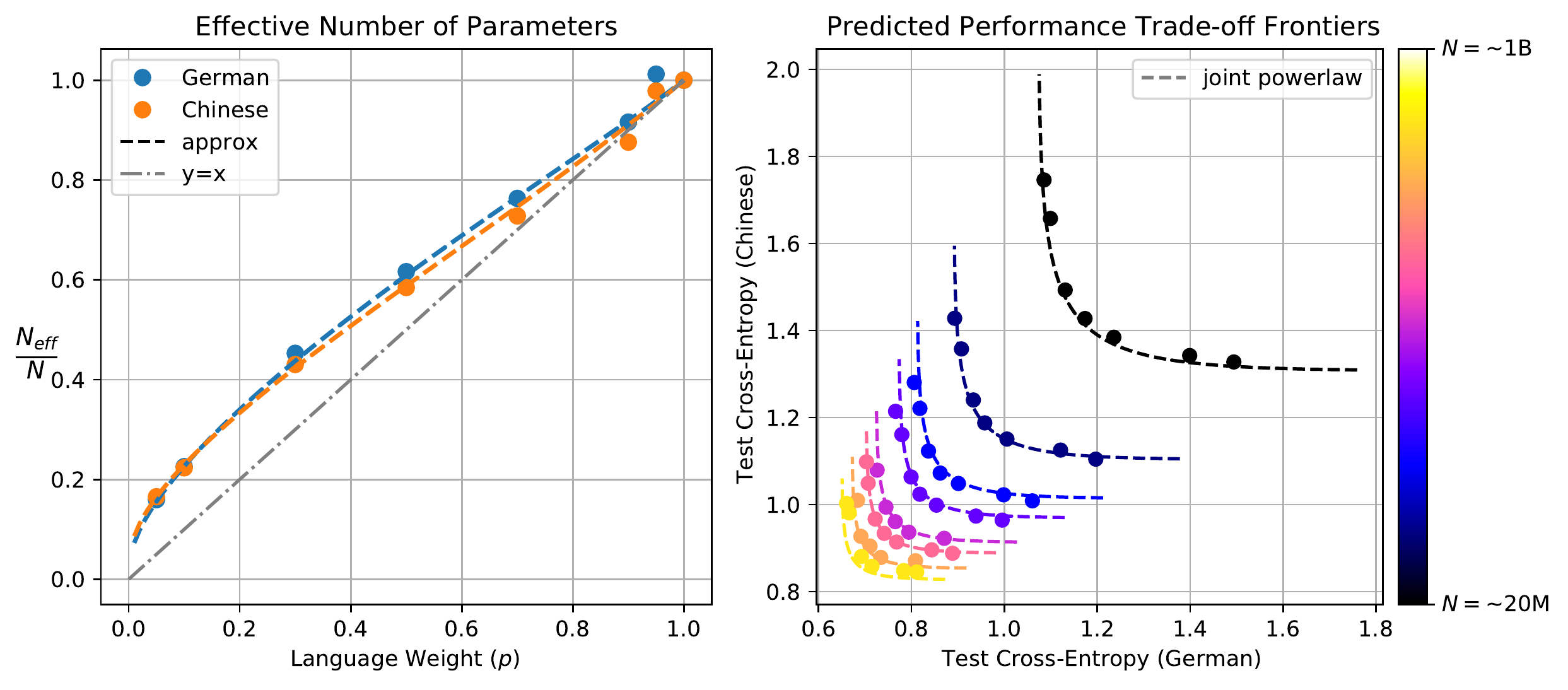}
\vspace{-0.2cm}
\caption{\label{fig:ratio_and_pareto} Approximate joint scaling laws described by equations (\ref{eq:estimated_f}) and (\ref{eq:f_hat}) almost perfectly capture the language pair interactions across all scales. \emph{Left:} The fitted approximation $\hat{f}$ described in \autoref{eq:f_hat}. \emph{Right:} The predicted performance trade-off frontier (dashed lines) as well as the empirically observed trade-off values.}
\vspace{-0.3cm}
\end{figure*}

In stark contrast to the En$\rightarrow$XX setting, when translating into English, we observe significant positive synergy among the language pairs. This observation aligns well with recent results in the literature showing multilingual models achieving SOTA performance for translation to English \citep{palm, lepikhin2020gshard}. It is unclear if this synergy arises as a specificity of having English as the target language or because multilingual encoding is intrinsically more amenable to parameter sharing than multilingual decoding. Understanding the exact dynamics giving rise to such positive interaction between the language pairs is an exciting open question. \looseness=-1

\textbf{Benefits for Massive Multilingual Models:} While we observed minimal synergy between En$\rightarrow$XX languages pairs, and therefor minimal gains in \textit{absolute} effective capacity, if we look at \textit{relative} effective capacity, we can see considerable benefits in using multilingual models for language pairs with small weight. For example, a model trained for En$\rightarrow$\{De, Zh\} with $5\%$ weight on German has an effective capacity of more than $3\times$ a model trained with $5\%$ capacity of this model for only German. These relative gains are even more evident when there is positive task synergy, such as for XX$\rightarrow$En, where models train with $5\%$ weight have more than $6\times$ gain in (effective) parameters. This hints that, if these findings generalize beyond the two-task setup\footnote{see \autoref{app:trilingual} for preliminary experiments on models trained on more than two language pairs.}, then training large multilingual models for training mixtures with a large number of small weight language pairs is significantly more memory efficient than training separate small models for each language pair. \looseness=-1

\subsection{Guiding Language Balancing} \label{subsec:task_balancing}

As discussed earlier, one of the areas where multilingual scaling laws can be most impactful is in guiding language balancing/weighting when training large multilingual models, an open problem that has been studied extensively \citep{Arivazhagan2019MassivelyMN, wang-etal-2020-balancing}. However, in its current form, our (joint) scaling law can only be use to decide between weightings that were used for fitting it and cannot be used to predict performance on new, unseen weightings, as $\beta_{p,i}$ needs to be estimated empirically.

To extend to unseen language pair weightings, we instead focus on estimating $f_i(\cdot)$. Given access to $f_i(p)$, accurate prediction of $\Ls_{i}(N)$ for \textbf{any weighting} can be achieved by using the \textbf{single-language-pair scaling law}:
\begin{align} \label{eq:estimated_f}
    \Ls_i(N; p) = \beta_{1, i} \big(\hat{f}_i(p) N \big)^{-\alpha_{i}} + L_{\infty}^{(i)}. 
\end{align}

As observed in Section \ref{subsec:capacity}, $f_i(p)$ has a number of desirable properties that makes it easy to estimate: (i) it is invariant to test set and languages, (ii) it is smooth and generally well-behaved. As such, one can achieve an accurate approximation of $f$ with just a few data points.

We utilize this methodology to estimate the full task performance trade-off frontier for En$\rightarrow$\{De, Zh\} models. For estimating $f_i(\cdot)$, we fit an approximate joint scaling law of the form \autoref{eq:estimated_f}, where $\hat{f}_i(\cdot)$ is parameterized as 
\begin{align} \label{eq:f_hat}
    \hat{f}_i(p) = p + c_1 p^{c_2} (1 - p)^{c_3}
\end{align}

with $c_1,c_2,c_3$ being fitted coefficients. Figure \ref{fig:ratio_and_pareto} demonstrates our results; our procedure is able to almost perfectly capture the full task performance frontier across a variety of model scales. With access to such accurate predictions of the performance frontier, a practitioner can precisely determine how to weigh the individual language pairs during training based on her preferences and target model size. 

We should note that the choice of function class to fit $f_i(\cdot)$ is highly dependent on the practitioner's computational budget. In our case, we prioritized accuracy and used a flexible function class of the form (\ref{eq:f_hat}) for fitting. Such flexibility comes with the cost of needing to compute more empirical values to reliably estimate $f(\cdot)$. In the scenarios with more limited computational budget, we have observed that even rudimentary linear approximations of $f$ are able to provide accurate representations of the performance frontier. See \autoref{app:other_approx} for examples.

\begin{figure*}[h]
\centering
\includegraphics[width=.8\textwidth]{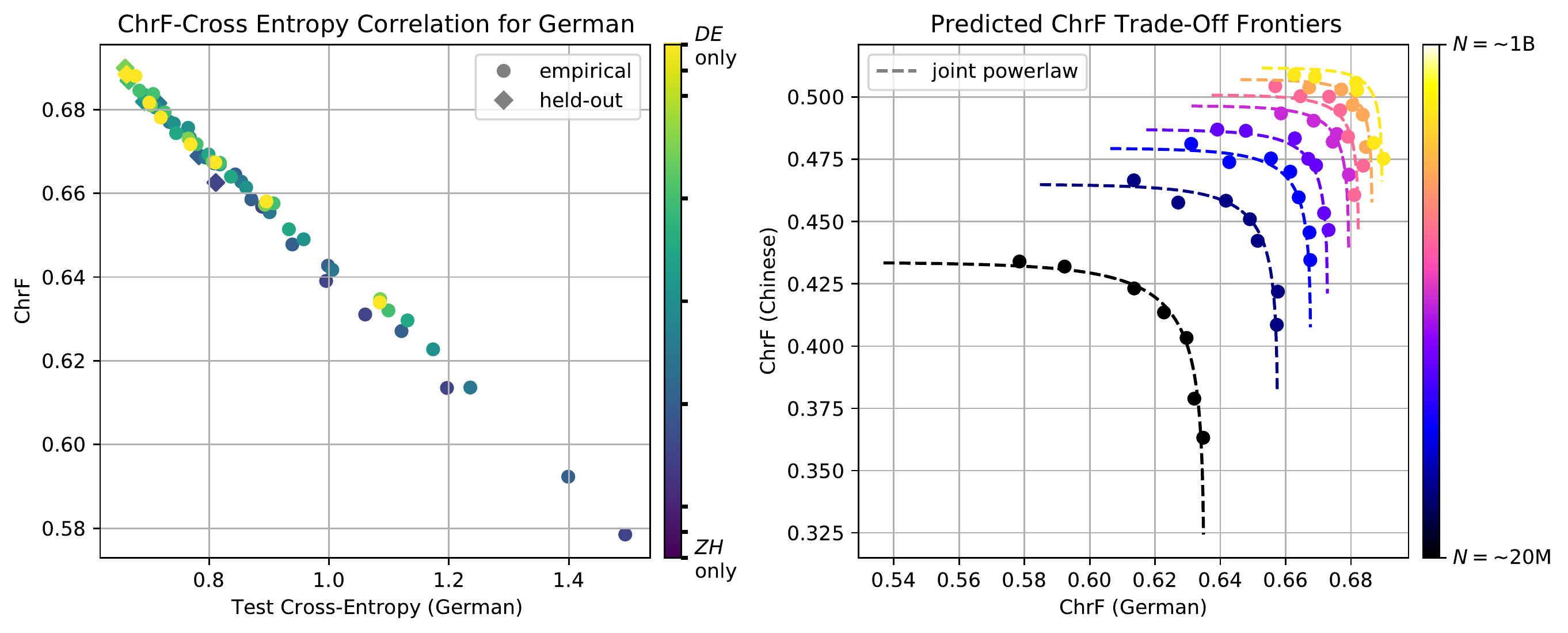}
\vspace{-0.25cm}
\caption{\label{fig:correlation_pareto_chrf} The generation quality behavior of our models as measured by ChrF. \emph{Left:} We observe consistent positive correlations between ChrF and cross-entropy loss. \emph{Right:} Our scaling laws can be used to generate accurate performance trade-off frontiers for ChrF.}
\vspace{-0.3cm}
\end{figure*}

\paragraph{Translation Quality} Finally, we note that in the MT literature, quality is often measured via metrics such as BLEU \citep{papineni-etal-2002-bleu}, ChrF \citep{popovic-2015-chrf} and BLEURT \citep{sellam-etal-2020-bleurt} as opposed to cross-entropy, since the latter doesn't account for the problem of \textit{decoding} translations from the models and is sometimes found to not correlate with human preferences \citep{koehn-knowles-2017-six}. As such, MT practitioners might be concerned regarding the applicability of these results for practical applications. To ensure that our findings also apply to the quality of translations, we decode translations from our trained models using beam search \citep{graves2012sequence} and evaluate how their quality changes as we scale the models, using ChrF and BLEURT.

\autoref{fig:correlation_pareto_chrf} (left) shows cross-entropy and ChrF scores for the En$\rightarrow$De language pair of our En$\rightarrow$\{De, Fr\} models, evaluated on the in-domain test set. We find that this automatic metric has an almost-linear relationship with cross-entropy, hinting that our observations also generalize from cross-entropy to generation quality. \autoref{fig:correlation_pareto_chrf} (right) also shows the predicted ChrF performance trade-off frontier obtained by fitting our joint scaling law (\autoref{eq:joint-scaling-law}) to the ChrF performance on the in-domain test set (parametrizing the effective parameter fraction function as in \autoref{eq:f_hat}). Our procedure is able to capture this trade-off frontier almost as well as the cross-entropy frontier. Similar findings for the BLEURT metric on out-of-distribution test sets can be found in \autoref{app:translation-quality}.

\section{Conclusions \& Future Work}

\label{sec:conclusion}
Current state-of-the-art large neural models are moving towards using as much data from as many domains, modalities and languages as possible to unlock exciting new capabilities. Unfortunately, a clear understanding of the behavior of such multitask models at scale is missing. This in turn slows down the model development process since practitioners have to resort to trial and error for balancing their tasks in their models.  In this paper, we attempted to take an initial step towards alleviating this problem by performing a large-scale study of the properties of multilingual models.

In particular, we attempted to study this problem from the lens of multilingual machine translation. We showed that, for each language pair and language pair weighting, a power-law describes the evolution of the model test performance as a function of the model size. We examined the dependence of the scaling law parameters on the language pair weights and demonstrated that the scaling exponent and the irreducible loss are independent of the weightings. Using these observations, we provided a novel joint scaling law that succinctly captures the scaling behavior across different model sizes and weightings and used it to define the notion of \textit{effective fraction of parameters} assigned to a language pair ($f_i(\cdot)$). We showed that this quantity robustly captures the language pair interactions and is surprisingly invariant to the similarity of the languages.  In the end, we sketched a procedure to use $f_i$ to estimate the task performance trade-off frontier for all model scales.

\paragraph{Future Work} In this paper, we studied the scaling behavior of multilingual translation models. Examining whether the conclusions of our work apply to multi-task setups beyond translation is a promising research direction. Most of our conceptual framework and experimental setup can easily be reused for this since there is little difference in the mathematical formulation of the optimization problem and it is likely that similar observations regarding the lack of transfer in data-rich scenarios will be found, as multilinguality can be considered an easier subset of the broader multitask learning challenge.

Furthermore, to keep our investigation tractable, we focused most of our experiments on the two-language-pairs scenario. However, we believe the presented results should be easily extendable to models trained with more languages (see \autoref{app:trilingual}). We leave such extensions to future work. 

Finally, to simplify the model scaling behavior, we focused our analysis on the data-rich setup. However, in many applications, at least some of the tasks are mid- or low-resource. Extending these results to such scenarios is an interesting future direction.
\bibliography{paper}
\bibliographystyle{icml2023}

\newpage
\appendix
\onecolumn
\section{Model Sizes and Hyperparameters}
\label{app:model-details}

\begin{table}[!htp]\centering
\scriptsize
 \setlength{\tabcolsep}{2pt}
\begin{tabular}{cccccccccc}\toprule
\textbf{Enc. Layers} &\textbf{Dec. Layers} &\textbf{Emb. Dim} &\textbf{\# Heads} &\textbf{Head Dim} &\textbf{MLP dim} &\textbf{Vocab Size} &\textbf{\# Parameters} &\textbf{Corrected \# Parameters} \\\midrule
2 &2 &512 &8 &64 &2048 &128k &149,953,024 &18,881,024 \\
3 &3 &768 &12 &64 &3072 &128k &260,322,816 &63,714,816 \\
6 &6 &768 &12 &64 &3072 &128k &324,035,328 &127,427,328 \\
9 &9 &768 &12 &64 &3072 &128k &387,747,840 &191,139,840 \\
9 &9 &1024 &16 &64 &4096 &128k &601,931,776 &339,787,776 \\
12 &12 &1024 &16 &64 &4096 &128k &715,193,344 &453,049,344 \\
12 &12 &1280 &16 &80 &5120 &128k &1,035,876,864 &707,869,184 \\
12 &12 &1536 &16 &96 &6144 &128k &1,412,528,128 &1,019,312,128 \\
\bottomrule
\end{tabular}
\end{table}

\section{Individual Scaling Laws Fits}
\label{app:individual-scaling-law-fits}
\subsection{Out-of-Domain}
\begin{figure}[h]
\centering
\includegraphics[width=1\textwidth]{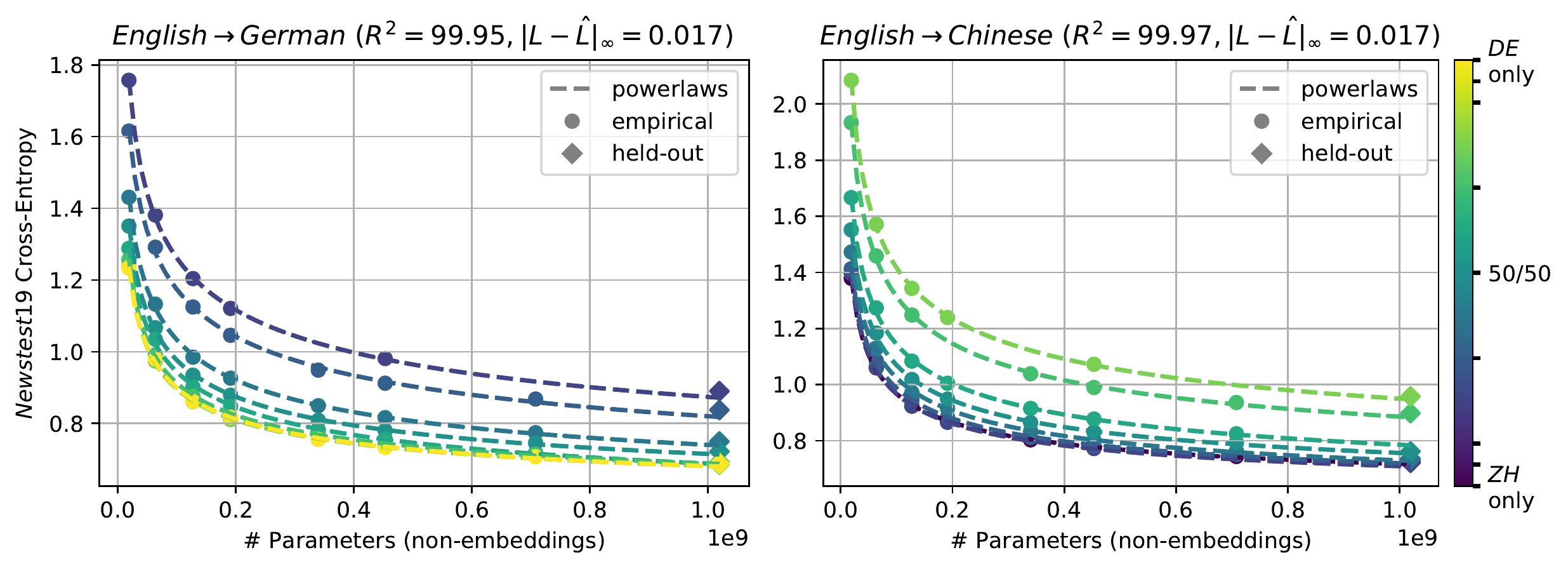}
\vspace{-0.6cm}
\caption{\label{fig:individual-sl-en-dezh_wmt_tgtorig} The evolution with model size of the cross-entropy loss on the \textit{newstest19} test set for En$\rightarrow$\{De, Fr\} models, as well as the fitted scaling laws. The color represents the weighting of the languages. Note that we don't show the \textit{zero-shot} behavior.}
\end{figure}

\begin{figure}[h]
\centering
\includegraphics[width=1\textwidth]{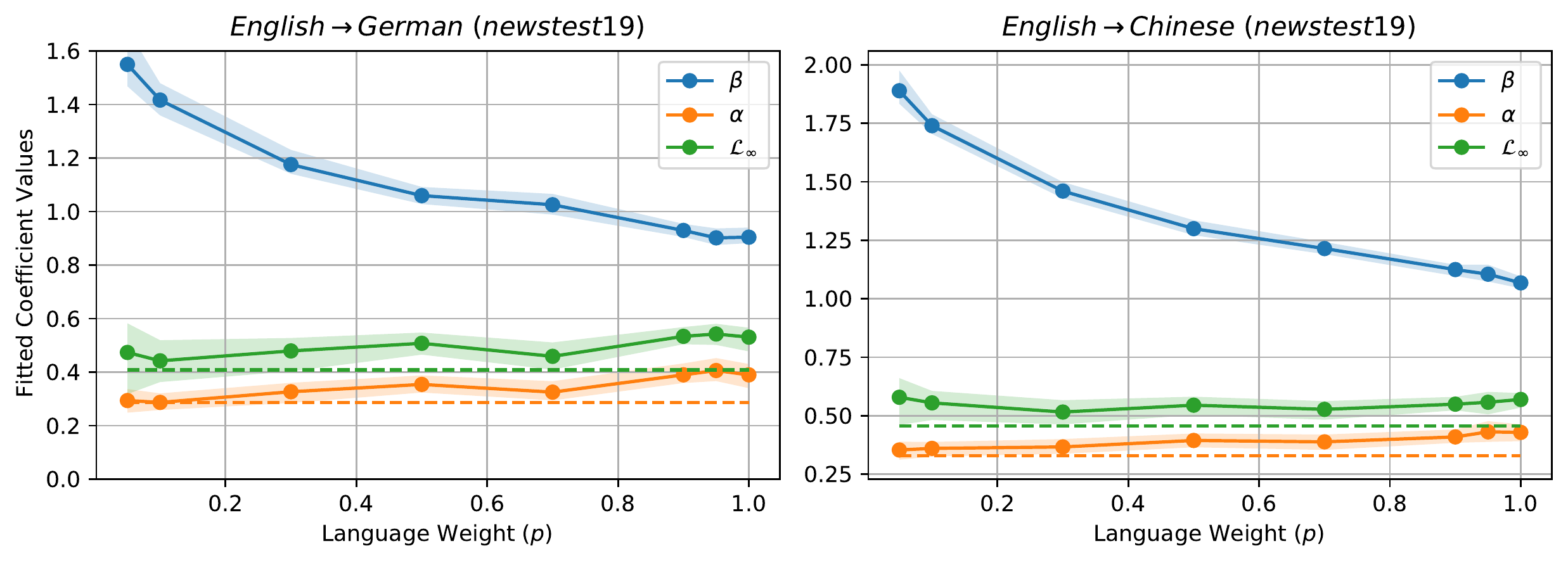}
\vspace{-0.5cm}
\caption{\label{fig:scaling_coefficients_en_dezh_wmt_tgtorig}Coefficient values, for scaling laws fitted on \textit{newstest2019}, for German (left) and French (right) as a function of the language weight, with the shaded region representing the standard deviation. The dashed lines represent the value of jointly fitted coefficients from \autoref{eq:joint-scaling-law}}
\end{figure}

\subsection{English$\rightarrow${German, French}}
\begin{figure}[h]
\centering
\includegraphics[width=1\textwidth]{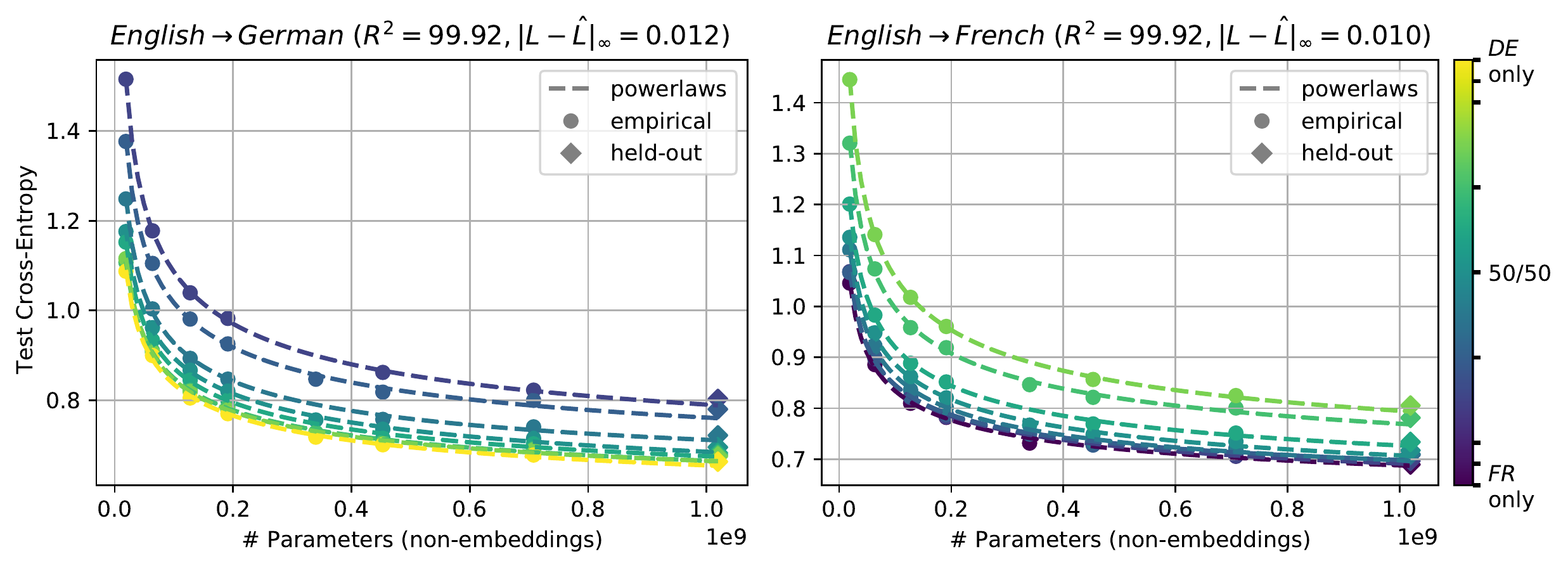}
\vspace{-0.6cm}
\caption{\label{fig:individual-sl-en-defr} The evolution of the (in-domain) test cross-entropy loss with model size for En$\rightarrow$\{De, Fr\} models, as well as the fitted scaling laws. The color represents the weighting of the languages. Note that we don't show the \textit{zero-shot} behavior.}
\end{figure}

\begin{figure}[h]
\centering
\includegraphics[width=1\textwidth]{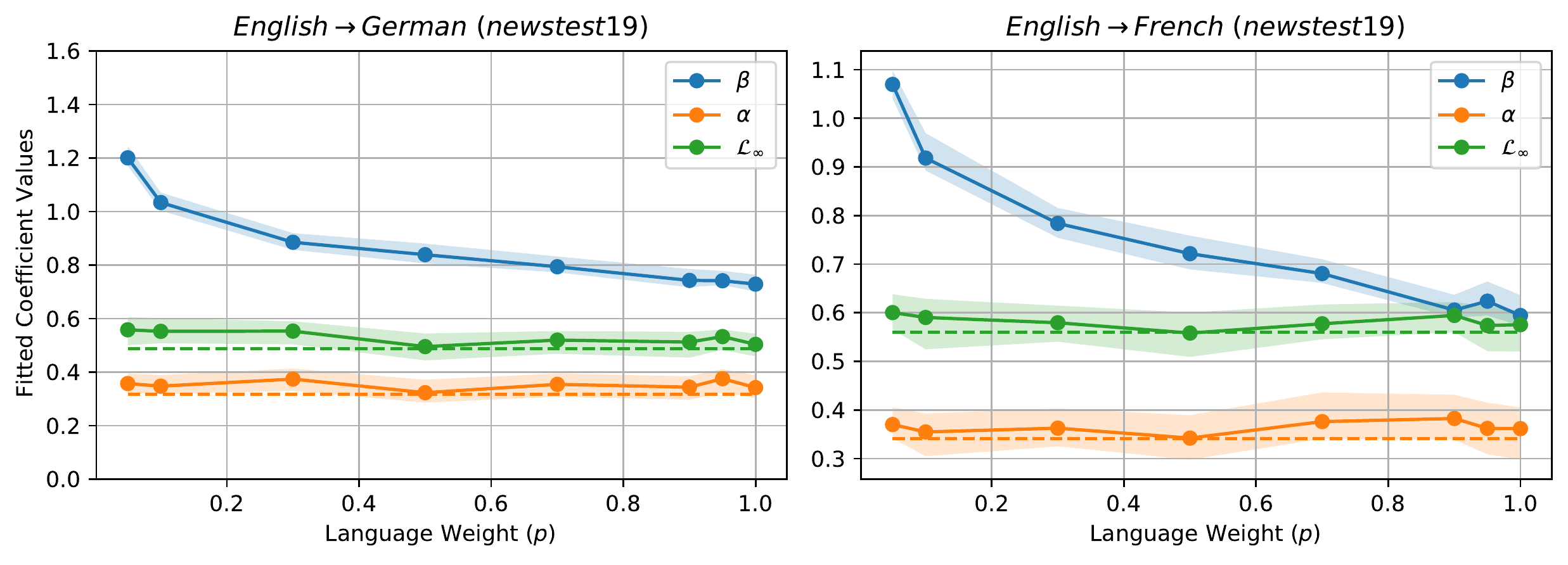}
\vspace{-0.5cm}
\caption{\label{fig:scaling_coefficients_en_defr}Coefficient values for German (left) and French (right) as a function of the language weight, with the shaded region representing the standard deviation. The dashed lines represent the value of jointly fitted coefficients from \autoref{eq:joint-scaling-law}}
\end{figure}

\subsection{{German, Chinese}$\rightarrow$English}

\begin{figure}[h]
\centering
\includegraphics[width=1\textwidth]{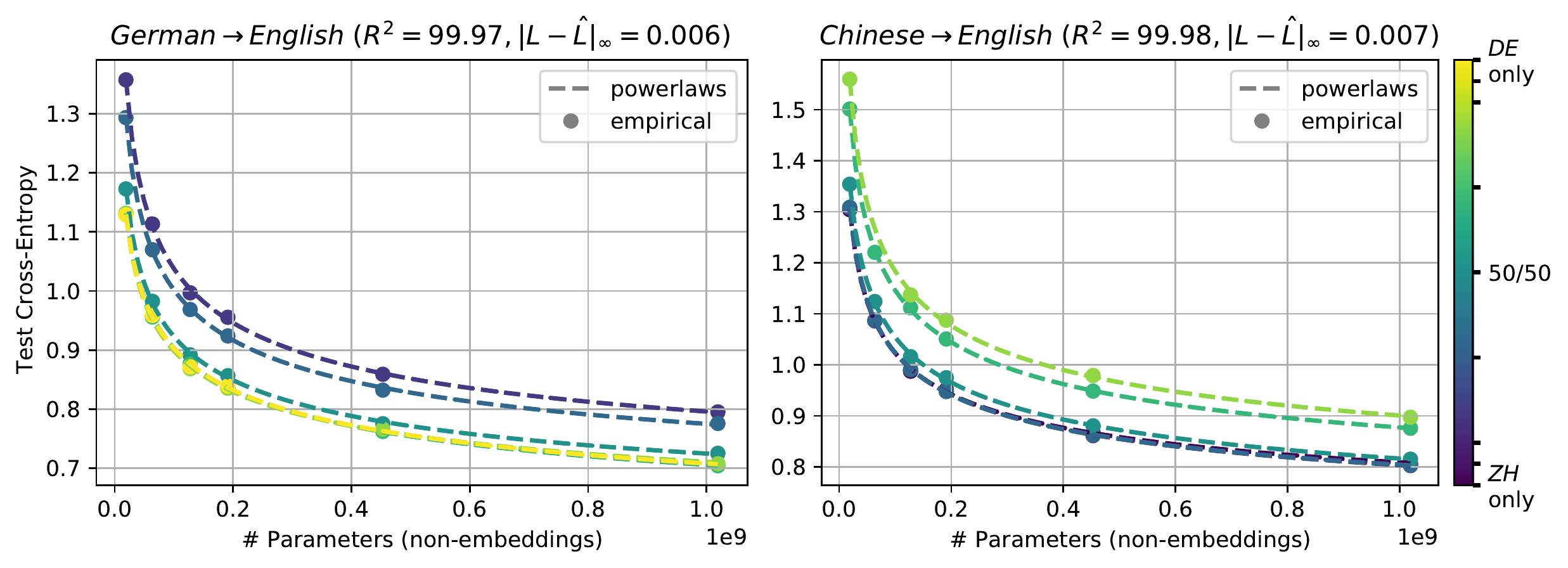}
\vspace{-0.6cm}
\caption{\label{fig:individual-sl-en-defr} The evolution of the (in-domain) test cross-entropy loss with model size for \{De, Zh\}$\rightarrow$En models, as well as the fitted scaling laws. The color represents the weighting of the languages. Note that we don't show the \textit{zero-shot} behavior.}
\end{figure}

\begin{figure}[h]
\centering
\includegraphics[width=1\textwidth]{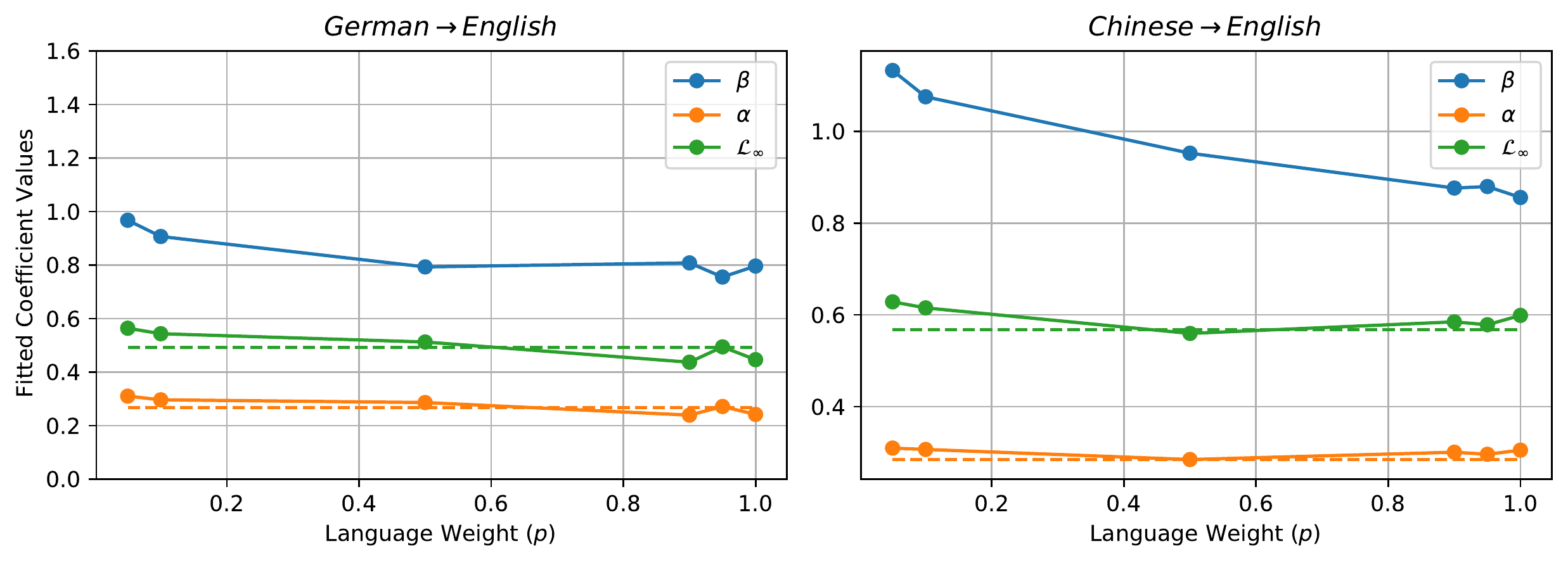}
\vspace{-0.5cm}
\caption{\label{fig:scaling_coefficients_en_defr}Coefficient values for German (left) and French (right) into English as a function of the language weight. The dashed lines represent the value of jointly fitted coefficients from \autoref{eq:joint-scaling-law}.We omit uncertainty estimates since less model capacities were used to fit the scaling laws, and therefor these estimates would be unreliable.}
\end{figure}

\section{Joint Scaling Law Fits}
\label{app:joint-scaling-law-fits}
\subsection{Out-of-Domain}
\label{app:joint-scaling-law-fits-ood}
\begin{figure}[h]
\centering
\includegraphics[width=1\textwidth]{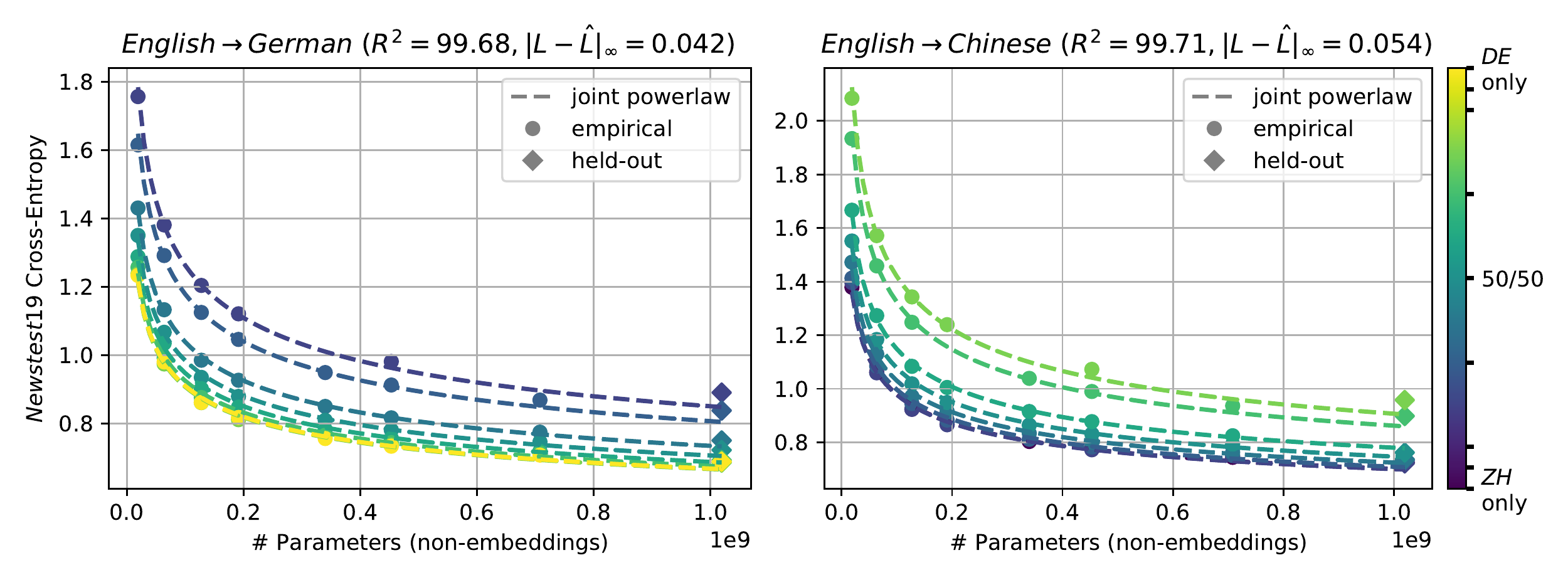}
\caption{\label{fig:joint-sl-dezh-en-wmt} The \textbf{joint} scaling law (\autoref{eq:joint-scaling-law}) fitted to models trained for En$\rightarrow$\{De, Zh\} models. Test loss here is evaluated on the \textit{newstest2019} test set.}
\end{figure}

\subsection{English$\rightarrow$\{German, French\}}
\label{app:joint-scaling-law-fits-en-defr}
\begin{figure}[h]
\centering
\includegraphics[width=1\textwidth]{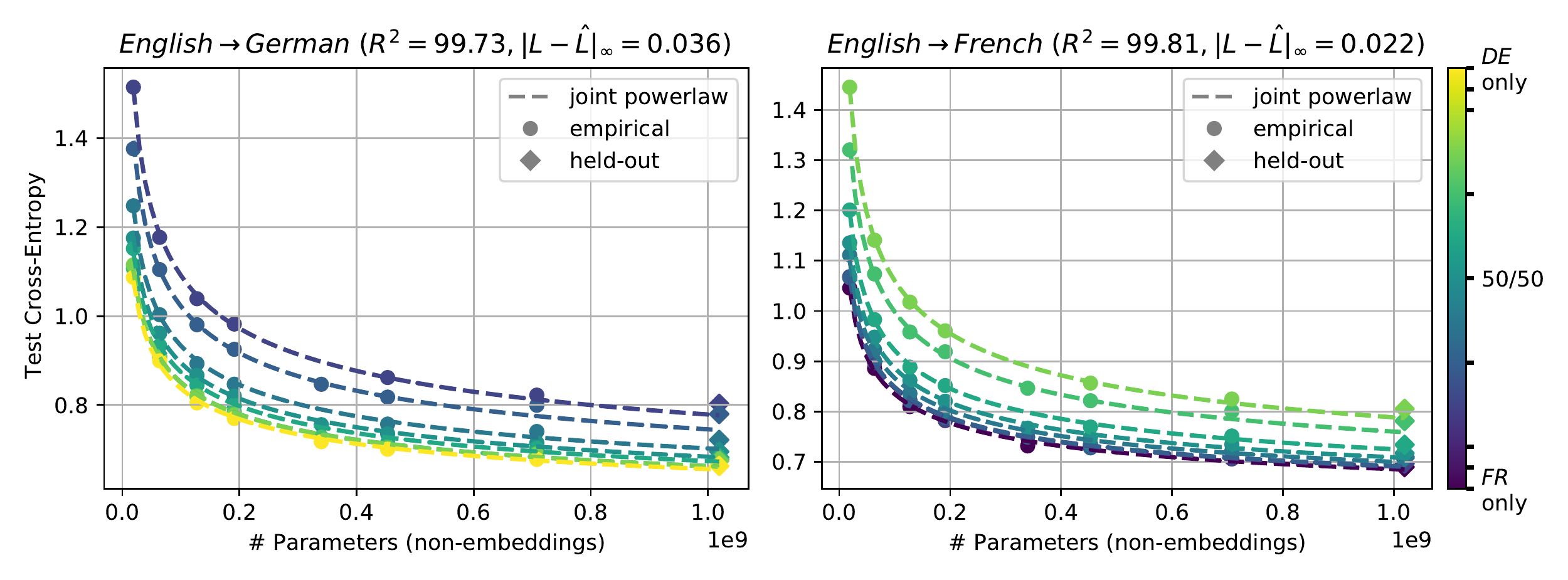}
\caption{\label{fig:joint-sl-en-defr} The \textbf{joint} scaling law (\autoref{eq:joint-scaling-law}) fitted to models trained for En$\rightarrow$\{De, Fr\} models. Test loss here is evaluated on in-domain test sets.}
\end{figure}
\subsection{\{German, Chinese\}$\rightarrow$English}
\label{app:joint-scaling-law-fits-dezh-en}
\begin{figure}[h]
\centering
\includegraphics[width=1\textwidth]{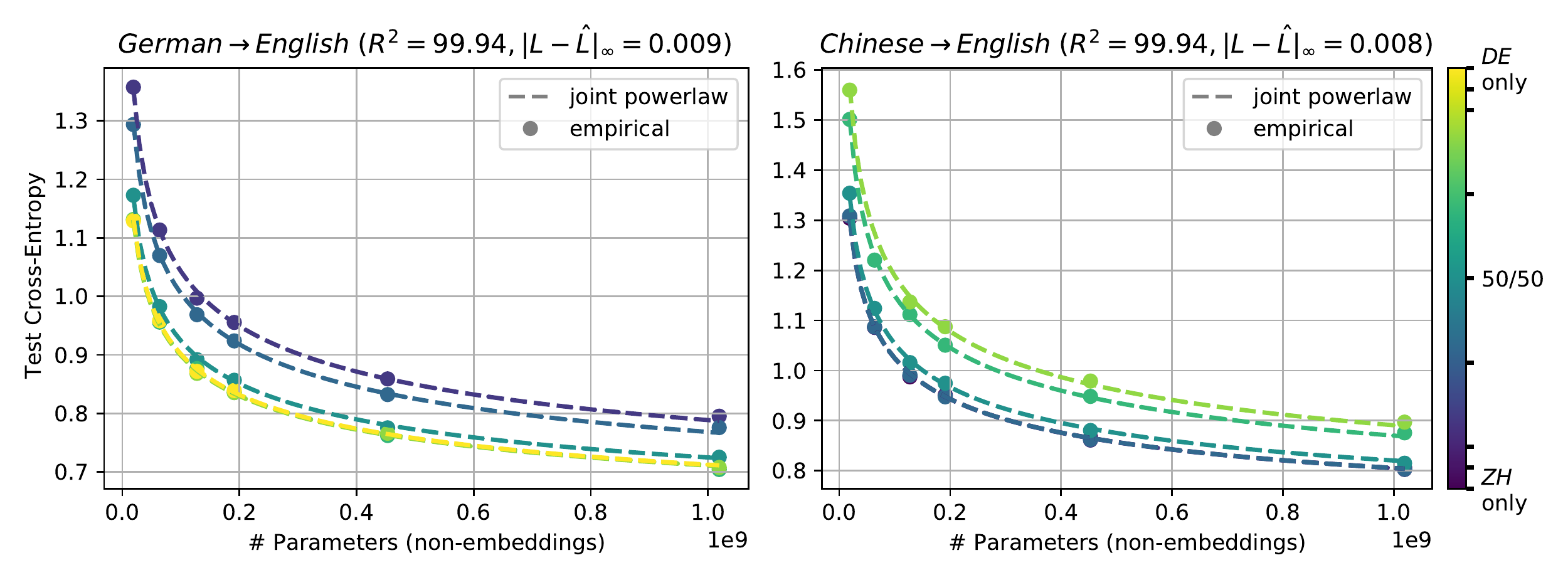}
\caption{\label{fig:joint-sl-dezh-en} The \textbf{joint} scaling law (\autoref{eq:joint-scaling-law}) fitted to models trained for \{De, Zh\}$\rightarrow$En models. Test loss here is evaluated on in-domain test sets.}
\end{figure}

\clearpage
\section{Derivation of the Effective Number of Parameters}
\label{app:model-ratio-derivation}
\begin{align*}
    \Ls_i(N; p) &= \beta_{p, i} N^{-\alpha_{i}} + L_{\infty}^{(i)} \\
                &= \beta_{1, i} \left (\frac{\beta_{p, i}}{\beta_{1, i}}\right) N^{-\alpha_{i}} + L_{\infty}^{(i)} \\
                &= \beta_{1, i} \left (\left (\frac{\beta_{p, i}}{\beta_{1, i}}\right )^{-\frac{1}{\alpha_i}} N \right )^{-\alpha_{i}} + L_{\infty}^{(i)} \\
                &= \beta_{1, i} \left (\left (\frac{\beta_{1, i}}{\beta_{p, i}}\right )^{\frac{1}{\alpha_i}} N \right )^{-\alpha_{i}} + L_{\infty}^{(i)} \\
                &= \beta_{1, i} N_\text{eff} ^{-\alpha_{i}} + L_{\infty}^{(i)} \\
                &= \Ls_i(N_\text{eff}; p)     
\end{align*} 

\section{Other Approximations to the Effective Parameter Ratio}
\label{app:other_approx}

We use a linear approximation of the form 
\begin{align} \label{eq:f_hat_linear}
    \hat{f}_i(p) = c_1 (p - 1) + 1.
\end{align}

\begin{figure}[h]
\centering
\includegraphics[width=1\textwidth]{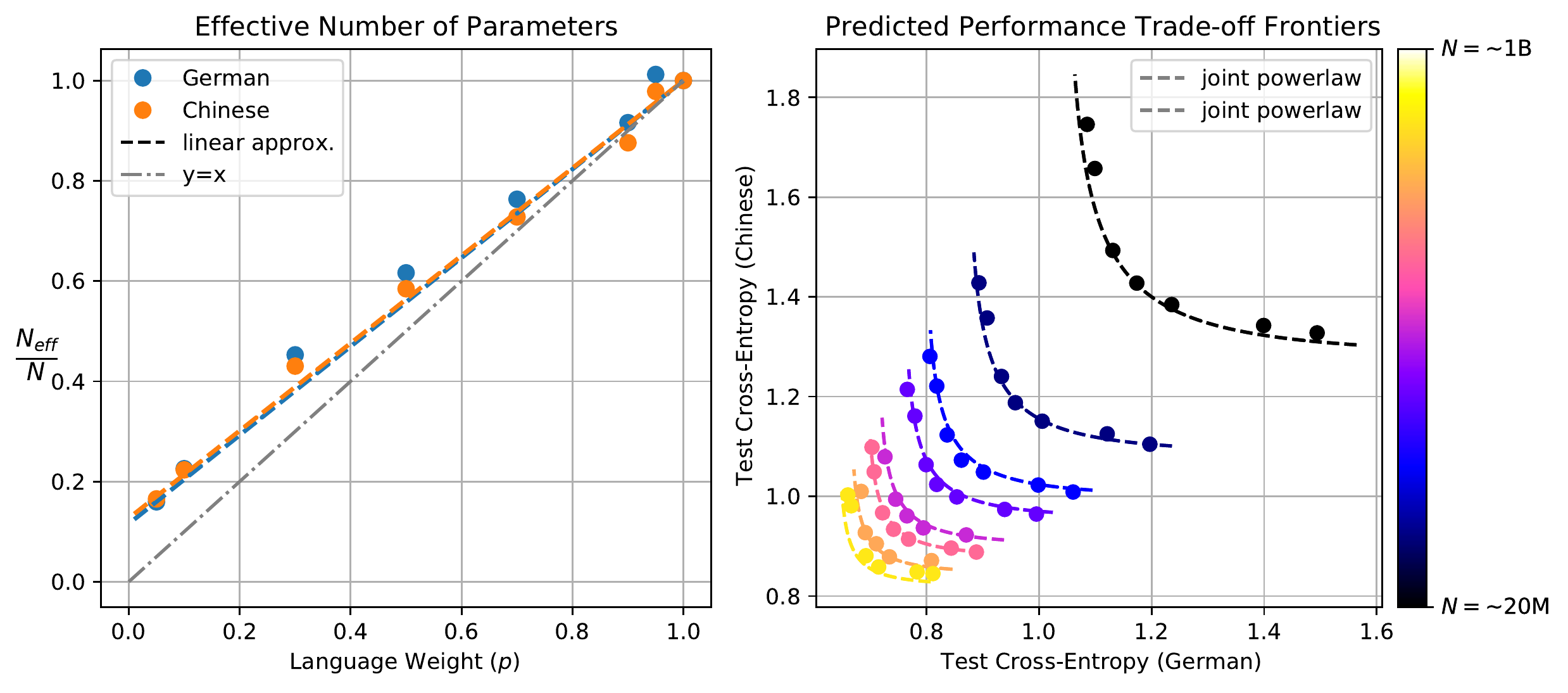}
\caption{\label{fig:ratio_paretto_linear} Approximate joint scaling laws described by equations (\ref{eq:estimated_f}) and (\ref{eq:f_hat_linear}) is able to capture the task interactions across all scales well, even with single fitted coefficient for ratio function. \emph{Left:} The fitted approximation $\hat{f}$ described in \autoref{eq:f_hat}. \emph{Right:} The predicted performance trade-off frontier (dashed lines) as well as the empirically observed trade-off values.}
\end{figure}

\clearpage
\section{Translation Quality}
\label{app:translation-quality}

\begin{figure}[h]
\centering
\includegraphics[width=1\textwidth]{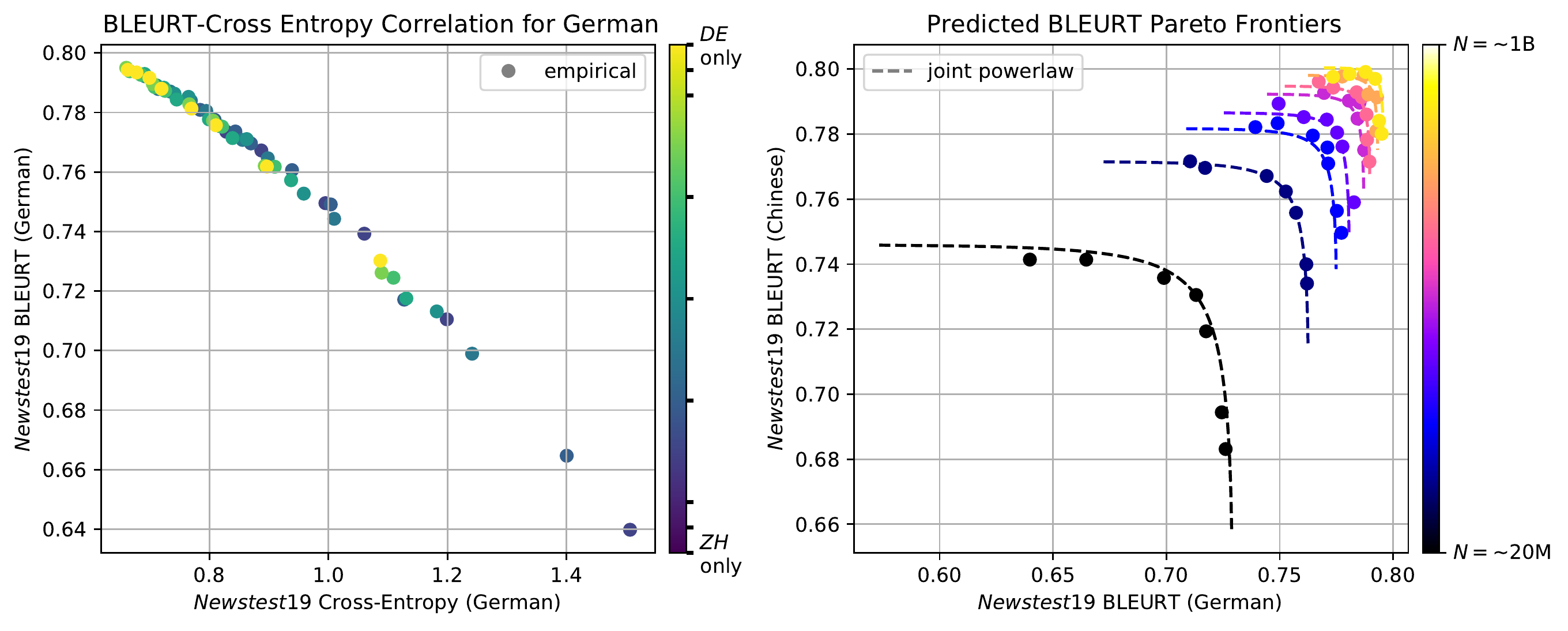}
\caption{\label{fig:correlation_pareto_bleurt_wmt} (left) shows cross-entropy and BLEURT scores for the En$\rightarrow$De language pair of our En$\rightarrow$\{De, Fr\} models, evaluated on the \textit{newstest19} test set. We find that this automatic metric has an almost-linear relationship with cross-entropy, hinting that our observations also generalize from cross-entropy to generation quality. \autoref{fig:correlation_pareto_chrf} (right) also shows the predicted BLEURT performance trade-off frontier obtained by fitting our joint scaling law (\autoref{eq:joint-scaling-law}) to the BLEURT performance on the \textit{newstest19} test set (parametrizing the effective parameter fraction function as in \autoref{eq:f_hat}).}
\end{figure}

\section{Convergence Correction}
\label{app:convergence-correction}
Due to \textit{implicit} scalarization, models trained with very little task weight ($<0.1$) will see less than a full epoch of that task's data, even when trained with 1M steps. I our experiments we saw that this was causing problems in the fit the scaling laws due to an \textit{undertraining} of our largest models. 

To mitigate this problem without training these models for a prohibitively large number of steps, we apply recent findings in learning curve \citep{learning-curve-theory} to estimate the performance of largest models trained with $p\leq0.05$ task weight at convergence, by fitting a power-law to the performance evolution as training progresses, and predicting the performance of these models at 2.5M steps. This only affect two models per scenario considered.

\section{Extension to more than two languages/tasks}
\label{app:trilingual}

As an early effort to understand if our findings apply to more than two tasks, we trained various model sizes for to translate into \textbf{three} languages (EEn$\rightarrow$\{De, Fr, Zh\}), and compared the predictions using the scaling laws for models trained on two language pairs (En$\rightarrow$\{De, Zh\} and En$\rightarrow$\{De, Fr\}). 

\begin{figure}[h]
\centering
\includegraphics[width=1\textwidth]{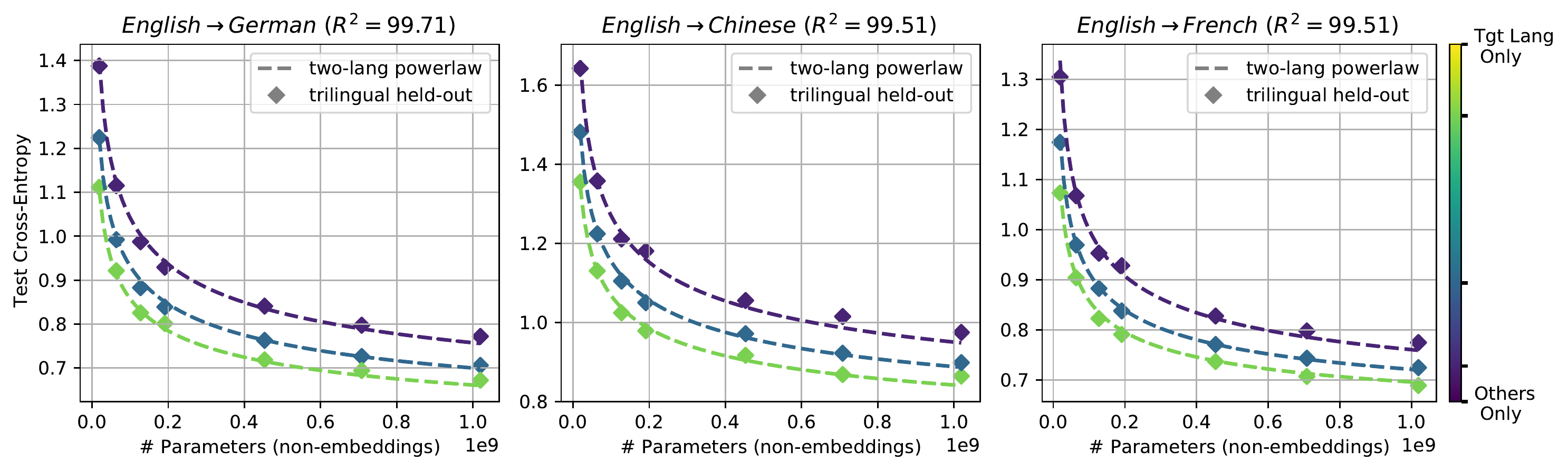}
\caption{\label{fig:trilingual-scaling-laws} The evolution of the (in-domain) test cross-entropy loss with model size for En$\rightarrow$\{De, Fr, Zh\} models, as well as the fitted scaling laws fitted for En$\rightarrow$\{De, Zh\} (left and middle) and En$\rightarrow$\{De, Fr\} (right). The color represents the weighting of the languages. Note that we don't show the \textit{zero-shot} behavior.}
\end{figure}

Figure \autoref{fig:trilingual-scaling-laws} shows the results. Overall we find that (combination of) the joint scaling laws fitted on models trained on two language pairs predict well the performance of models trained for three language pairs, showing that the invariances found in previous sections generalize to settings with more than two tasks. These results also hint that computation of effective parameters counts for multi-task models with many tasks can be simplified and made more tractable by training models with much smaller subset of tasks.  

\end{document}

%% file: math_commands.tex
\usepackage{amsmath,amsfonts,bm}

\def\eqref#1{equation~\ref{#1}}

\def\1{\bm{1}}

\def\eps{{\epsilon}}

\def\vtheta{{\bm{\theta}}}

\def\vw{{\bm{w}}}

\DeclareMathAlphabet{\mathsfit}{\encodingdefault}{\sfdefault}{m}{sl}
\SetMathAlphabet{\mathsfit}{bold}{\encodingdefault}{\sfdefault}{bx}{n}

\newcommand{\Ls}{\mathcal{L}}
\newcommand{\R}{\mathbb{R}}

\DeclareMathOperator*{\argmin}{arg\,min}